# Enabling energy efficient machine learning on a Ultra-Low-Power vision sensor for IoT

Francesco Paissan[1], Massimo Gottardi[2], Elisabetta Farella[1]
[1]*E3DA - ICT-irst,* [2]*IRIS - CMM*
*Fondazione Bruno Kessler*
Trento, Italy
{fpaissan, gottardi, efarella}@fbk.eu

*Abstract*—The Internet of Things (IoT) and smart city paradigm includes ubiquitous technology to extract context information in order to return useful services to users and citizens. An essential role in this scenario is often played by computer vision applications, requiring the acquisition of images from specific devices. The need for high-end cameras often penalizes this process since they are power-hungry and ask for high computational resources to be processed. Thus, the availability of novel low-power vision sensors, implementing advanced features like in-hardware motion detection, is crucial for computer vision in the IoT domain. Unfortunately, to be highly energy-efficient, these sensors might worsen the perception performance (e.g., resolution, frame rate, color). Therefore, domain-specific pipelines are usually delivered in order to exploit the full potential of these cameras. This paper presents the development, analysis, and embedded implementation of a real-time detection, classification and tracking pipeline able to exploit the full potential of background filtering Smart Vision Sensors (SVS). The power consumption obtained for the inference - which requires 8ms - is 7.5 mW.

## I. Introduction

Vision Sensor Nodes (VSN) are the building blocks of Wireless Vision Sensors Networks (WVSN), a promising technology for its extended application domain that ranges from structural health analysis [1] to urban scenarios observation [2] [3]. While developing an algorithm for a VSN, it is crucial to consider the amount of data transmitted, which can be prohibitive [4] and have a high impact on energy efficiency. Literature suggests two viable solutions: bandwidth compression [5] or hardware solutions such as the near-sensor image processing paradigm [6] [7]. For the latter, it is crucial to reduce the amount of redundant information to be processed; therefore, event-based Smart Vision Sensors (SVS) enabling in-hardware motion detection algorithms are used [8] [9].

To exploit the full potential of these sensor technologies - which usually are also characterised by lower quality images (e.g. lower spatial and pixel resolution) - application-specific processing pipelines are the most promising option, offering a trade-off between performance and power consumption [10] [11] [12].

To prove the efficiency of these pipelines, we implemented a detection, classification and tracking pipeline on images from a low-power event-based VGA sensor with background subtraction and basic analysis capabilities. Therefore, the contributions of the work are: in

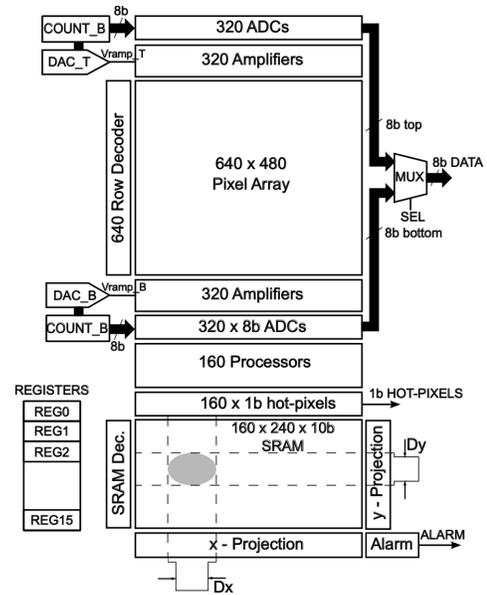

Fig. 1: Block diagram of the vision sensor generating an alert signal in presence of a moving target.

- the design of a sub-optimal detector targeting low operations count;
- the design of a pipeline targeting resource-constrained platforms for the IoT, implementing detection, classification and tracking of cars and pedestrians;
- MCU implementation and power consumption analysis of the proposed approach.

## II. Related Work

Camera-based solutions for object detection already exist but, as already mentioned, are costly from a computational and energy point of view [13]. In recent years, research focused on investigating alternatives to GP-GPUs clusters to fit resource-constrained platforms for applications in robotics, smart maintenance, and Internet of Things (IoT) [14] [15]. However, many of these applications use off-the-shelf processing units that still have a power demand around $5\,W$, more than what a solar-powered node can harvest. This system's high power consumption is to be attributed to the computational power needed to analyze high-resolution image streams. Smart Vision

Sensors aim at reducing the amount of redundant information in images, thus enabling more efficient processing of the scene. Technically, the most popular approach to detect and remove redundant information before delivering it to the processing unit is frame difference. In [9], the authors implemented an advanced frame difference technique that exploits temporal averaging to augment the range of detectable objects. Also [16], [17] and [18] exploit the frame difference paradigm with some minor adaptations (e.g. dynamic thresholding). More advanced techniques as the one implemented in [19], which exploits a 64x64 CMOS image sensor with on-chip moving object detection and rough localization, lack flexibility since the cluster for frame-recognition cannot be changed. One main drawback of this frame difference-based SVSs is that an event might be triggered in the presence of small movements of non-static objects as tree branches or sea waves.These are not useful targets for the proposed application, thus leading to higher power consumption.

Given the many advantages in exploiting the near-sensor processing paradigm, many studies on IoT-compatible processing pipelines as developed both for binary and RGB images. Binary image processing were of high interest in the last decades [20] [21] and afterward abandoned due to the improvements in the imaging and processing field. Nonetheless, this algorithms were rediscovered for the processing of ultra-low-power smart visual sensors generated images, where the binary output enables smart strategies to reduce power consumption and computational cost. [10] present a image processing node that implements Binarized Neural Networks (BNNs) to process the output of a SVS more efficiently. In [22] the authors exploit the synergy between smart sensors and energy-efficient quad-core cluster processor for data analysis. The evaluation setup is indoor and therefore it does not exploit the potential of SVSs, like the robustness to small movements, typical of outdoor environments. Also in [11] the authors exploit BNNs to filter pedestrian-related events, thus reasoning on higher order information for triggering event alarms.

## III. VISION SENSOR

In this work, we used a low-power VGA vision sensor embedding motion detection based on dynamic background subtraction. As shown in Fig. 1, the algorithm is applied on a QQVGA sub-sampled image, generating a motion bitmap of 120 x 160 pixel, where asserted pixels are called Hot-Pixels (HP). The bitmap is de-noised through a programmable erosion filter and used to generate two projection vectors along the *x* (160 pixels) and *y* (120 pixels) axes of the array. The size and aspect ratio of the moving target to generate the alert are assigned by setting constraints on the number of contiguous motion pixels in the (x,y)-projection vectors. The vision sensor operates in Motion Detection (MD), looking for events with no off-chip data delivering, and Imaging Mode (IM), activated after the alert, delivering VGA gray-scale image together with the related motion bitmap. A low-power FPGA controls the sensor and generates the interface with the external processor. After an event is detected, the sensor sets the ALARM to trigger the processor and starts delivering 10 consecutive frames, sufficient to execute visual tasks such as detection, classification and tracking of the objects that generated the alarm. The frames are delivered as gray-level VGA, together with the sub-sampled motion bitmap encoded into the least significant bit of the VGA pixels.
When available, the frames are delivered from the FPGA to a low-power MCU through a DCMI bus.

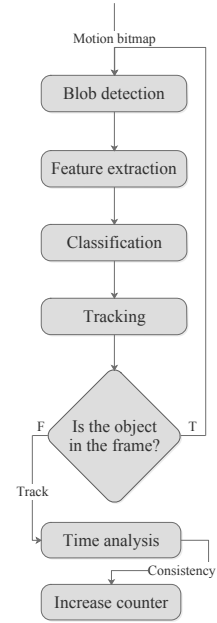

Fig. 2: Detection, classification and tracking pipeline.

Because the sensor is still in a prototyping stage, we are forced to sub-sample the images from the gray-scale VGA output, therefore adding overhead to the processing stage caused by the acquisition of a 4x bigger frame and the extraction of the motion information. As soon as an application-specific interface for the sensor will be ready, this operation will be avoided and the MCU would receive as independent signals the (x,y)-projections of the alarm-triggering object, therefore removing even more pre-processing from the vision pipeline.

## IV. PROCESSING PIPELINE

We implemented a pipeline for detection, classification and tracking of pedestrians and cars. For this, we exploited the features of the SVS described in the previous section. Our work is to be considered as an extension of the pipeline proposed for detection and classification in [23].

We refer to Figure 2 for an illustration of the algorithm presented in this work.

### A. Detection

Object detection - as intended in the computer vision domain - is the task of extracting instances of semantic objects (belonging to a certain class) from images. Our detection algorithm is developed for binary images, encoded as 1-pixels (active pixels) when movement is detected and 0-pixels (inactive pixels) where no movement is detected.

Formally, to detect regions of interest (ROI) from the (x,y)-projections, we consider the falling and rising edges of the projections and compute the Cartesian product between the rising and falling edges on the two axes obtaining the upper left and lower right corners as output sets respectively. Therefore, without considering the temporary sensor's limitations, the only operations needed to run the detection stage are memory read and write, negligible with respect to operations needed to run Suzuki's algorithm. Afterwards, in the feature extraction stage, the moments of the extracted distributions are computed and used to remove empty regions from the proposed set.

Despite the significant improvement in operation count for detection, this approach does not seem to impact classification and tracking performance on average. On the other hand, in some complicated scenarios with complex binary representation 3 this approach would inevitably perform worst than Suzuki's method. But those scenarios are not characteristic of the proposed application and, given the low-resolution of the sensor, are not likely to be detectable if present.

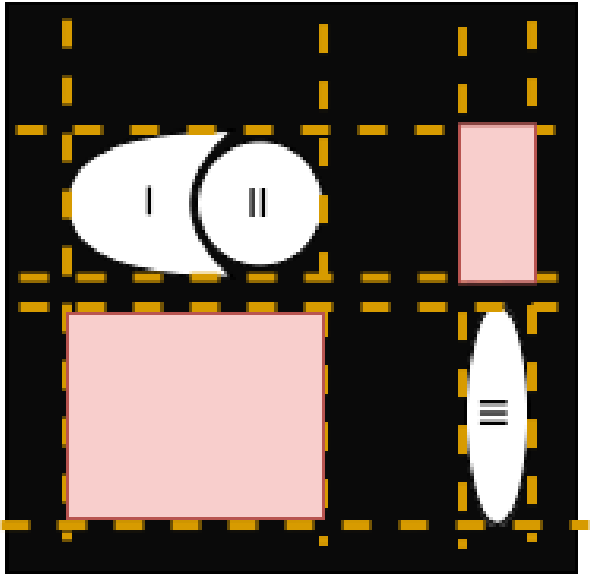

Fig. 3: Downsides on the proposed region proposal algorithm. Red rectangles are extra area which will be removed in the feature extraction process. Blob II is not visible.

As regard the detection sub-task, the contribution of this work is the development of a highly specific algorithm, which aims at being as neutral as possible from a computational cost analysis with respect to the rest of the pipeline. Our approach is compared with the method implemented in [23] and proposed by [21], which is highly accurate but has a significant computational cost – almost 80% of the required operations for the detection and classification pipeline are due to the choice of the detection algorithm.

The proposed detection algorithm is a two-stage detector. In particular, we developed a region-proposal algorithm that consists of a Cartesian product and aims at reconstructing the regions that triggered the alarm from the (x,y)-projection. From Figure 3, it is clear that this approach is limited with respect to the more sophisticated approach developed by Suzuki et al. [21] in extracting higher-level information (e.g., the hierarchical structure of contours) and less false positives. In fact, using our approach the system is more likely to detect false positives and therefore a filter for void pixel areas in needed. However, given the sensor's application domain and considering the SVS output, those limitations are not a big concern. In fact, this algorithm was tested on a sequence on 800 frames from an urban scenario recorded using the SVS. The results showed that, on average, our region proposal algorithm detects one extra object with respect to Suzuki's algorithm. Moreover, we computed the intersection-over-union of the non-empty bounding boxes with the ones detected by Suzuki's algorithm and found that, on average, it is around 0.30. This is Because the region-proposal algorithm detects bounding boxes which area is around 1.5 times compared with the more sophisticated method.

## B. Classification

For the classification stage, we validate the model proposed in [23]. In fact, classification is based on a linear kernel Support Vector Machine (SVM), trained on an artificially created dataset with 264 instances equally distributed between the two classes. In fact, we compared classification results using different features as for example invariant moments. Finally, we concluded that the accuracy of the model did not change much and remains nearly 98% on the artificial dataset. To add value to the proposed solution, which is already good from a computational perspective, we run tests to understand how the model behaves with respect to the features, which are:

- its area;
- the variances of the distribution associated to it on both axes, which are considered as descriptors of its shape;
- the minimum vertical coordinate of the bounding box including each blob, due to its correlation with respect to the scaling of the area.

As expected from a visual analysis of the sensor's output, we can see from Table I that the permutation importance [24] of the variance on the y axis is the most important features for the classification, together with the blob's area.

| Feature | Weigth |
|---|---|
| $\sigma_y^2$ | 0.38 ±0.05 |
| Area | 0.16 ±0.04 |
| Blob position | 0.11 ±0.03 |
| $\sigma_x^2$ | 0.06 ±0.03 |

TABLE I: Permutation importance of classification features.

To better understand how the model behaves with respect to the two most important features, we used a second order Accumulated Local Effects (ALE) [25] plot (Fig. 4). As we can see by analysing the extracted plot, the model prediction is mainly static all over the feature space, except in the lower-right corner. Meaning that the model output varies more in that section of the feature space and therefore the predicted class changes presumably in that space.

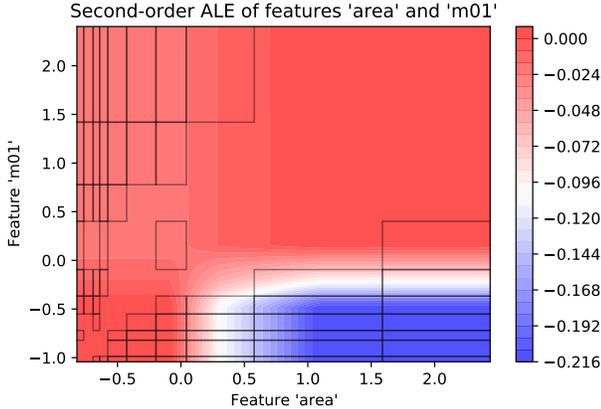

Fig. 4: Second order Accumulate Local Effects plot of features area and $\sigma_y^2$ (m01) already normalized. ALE computed on 10x10 bins.

## C. Tracking

The tracking algorithm implemented is inspired on the Simple Online Real-Time Tracking (SORT) algorithm proposed by Bewley et. al. in [26]. Particularly, the main structure is preserved and its independent blocks are modified to be more energy and computationally efficient.

*1) Detection:* This is the first stage of the tracking algorithm. In our implementation, we exploit the computational and energy efficiency of the algorithm already described.

*2) Estimation:* In this stage, a linear velocity model of the tracked object is updated during each observation of the object taken into account. This is performed using an implementation of Kalman filter [27], that exploits Single Instruction Multiple Data (SIMD). Despite we are aware that this is not the best possible representation of the object's dynamics, we observed during experiments that it works well enough to be used for our applications.

*3) Association:* This step is the core of the tracking algorithm, and consists in associating the same object in two consecutive frames. For this, many algorithms and methodologies might be involved [28], [29], but the most commonly used is the Munkres assignment algorithm (also known as the Hungarian algorithm), which proved to work well for temporary occlusions caused by moving objects.

The correlations between each blob $B_{t-1}$ in the frame $\Sigma_{t-1}$ (at time $t-1$) and each blob $B_t$ in the frame $\Sigma_t$ (at time $t$) can be represented like a bipartite graph. Specifically, the connected components in $\Sigma_{t-1}$ make up one part of the graph, and the connected components in $\Sigma_t$ make up the other part. By definition, the parts of the graph are fully connected and each edge has a weight computed as the overlap index (or Intersection over Union distance - IoU distance) between the detection boxes, including the blobs represented by the two connected nodes. Mathematically, for all $B_{t-1} \in O_{t-1}, B_t \in O_t$ the IoU distance, and therefore the weight of the edge from $B_{t-1}$ to $B_t$ is computed as:

$$IoU(B_{t-1}, B_t) = \frac{B_{t-1} \cap B_t}{B_{t-1} \cup B_t} \quad (1)$$

After defining the above structure, we solve for the optimal association using the Hungarian algorithm.

*4) Create or delete track:* In this stage, we insert or remove objects from the active objects list.
Every time a connected component enters the frame, if it is detected for more than $N_{hits}$ times it is considered as active and its state estimator is initialized using the last geometry recorded. Instead, if an already-tracked object is not detected for more than $T_{lost}$ frames, it is deleted from the active blob list.

*5) Time consistency and Counter:* For each tracked object, we want the algorithm to be reliable in all its stages. For this, we embed on each track a label referring to the class in which it was predicted the most of the time. This is based on the assumption that the classification might be wrong for about 10% of the samples, but should be reliable for the rest of the cases. Finally, after the object exits the frame, the consistency analysis in time is performed and the counter for the specific class is increased.

## V. RESULTS

In order to validate the presented pipeline, we performed the analysis described in this section.

### A. Validation sequences

Three sequences, for a total of 400 frames[1], were recorded at 8 fps using the development kit located at 7 meters above the ground and with a vertical orientation of $50°$. Since we wanted to validate the system at a broader level, we tried our algorithm also on a sequence taken from the PETS2009 dataset [30]. From this dataset, we selected a video in line with the aforementioned setup and we slightly adapted it in order to make it compliant with the sensor's simulator [31].

The labeling on all sequences was done manually by counting the number of cars and pedestrian appearing in the selected sequence.

### B. Tracking Performance

We use the detection and classification pipeline described, that on our sequences has an average accuracy of 85%. Specifically, for an instance to be correctly predicted in this analysis, we require that the predicted class matches the label, and that the bounding box has an IoU above 0.5.

In order to quantitatively evaluate the performance of the tracking algorithm, we computed the error for each class ($i$) as:

$$E^i = \frac{N_t^i - N_p^i}{N_t^i} \quad (2)$$

---
[1]The authors will keep an updated version of the dataset at https://github.com/fpaissan/SVS_BS-dataset.

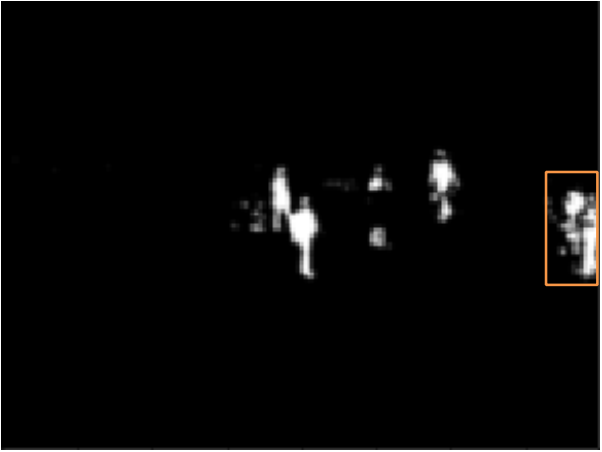

Fig. 5: Sensor output. In the orange rectangle there are two pedestrians.

where $N_t^i$ is the number of real objects belonging to class $i$ that passed in the frame, and $N_t^i$ is the number of tracked objects, always belonging to $i$. From definition, it follows that the error is zero when the tracking matches the label, negative if there are more tracked objects than labeled and positive otherwise.

TABLE II: Tracking performance over classes

|  | Ground Truth | | Tracking | |
| --- | --- | --- | --- | --- |
|  | Car | Pedestrian | Car | Pedestrian |
| Sequence 1 | 4 | - | 4 | - |
| Sequence 2 | 1 | 2 | 1 | 2 |
| PETS2009 Seq | - | 6 | - | 5 |

As shown in Table II, the performance of the pipeline (run with $T_{lost} = 1, N_{hits} = 6$) is good, despite it has a limited set of suitable application domains. In fact, for how it is developed, it does not perform well whenever two objects are close to each other. In fact, in this case it is possible that one of the two might be considered noise - see Figure 5. Despite this, we have an average accuracy of 94.6%.

This result is interesting, since it is obtained despite the detection and classification accuracy is around 85%. Moreover, this outcome implies that by pairing the detection and classification pipeline with the tracking algorithm, we can use less accurate algorithms for detection and classification. This can help in further reducing the power consumption of the system, broadening the algorithm choice, but of course always keeping in mind that the tracking is based on the detection result. Therefore, there must be a lower bound threshold for the detection accuracy, which permits the system to work.

*C. Power consumption*

To evaluate the power consumption of the proposed approach, we implemented the whole pipeline on a Cortex-M4 based board running at $80\,\text{MHz}$. The power consumption of the processing, without considering the overhead caused by the sensor is of $7.5\,\text{mW}$. Since the whole pre-processing due to the sensor limitation is condensed in the acquisition process, we expect a high power consumption for the acquisition stage. In fact, the measured consumption for reading 10 gray-scale frames and extract the motion bitmap is $60\,\text{mW}$. However, considering smaller images with lower bit resolution (from 8-bit gray-scale to 1-bit motion bitmap) and the event-triggered nature of the system, we conclude that it can be easily powered through energy harvesting techniques (e.g. by commercial off-the-shelf solar panels).

## VI. Conclusion

In conclusion, we presented a prototype board ready to be integrated in the Internet of Things as edge vision processing unit. Targeting resource-constrained platforms, we implemented a pipeline for the detection, classification and tracking of objects, which accuracy is around 95% and real-time implementation on a microcontroller consumes $7.5\,\text{mW}$ with only $8\,\text{ms}$ processing window. This is achieved mainly by exploiting the SVS capabilities and adapting the algorithm to the application specific input. This is a first step in our research on bringing the computation at the very edge, while using energy efficient but low-resolution vision sensors. Many aspects of the presented approach can be extended with the idea of exploiting more sophisticated hardware and software architectures, such as Convolutional Neural Networks (CNNs) and self-attention layers. In fact, we are working on an depth-wise separable convolution-based network [32] architecture based on YOLOv2 [33] able to scale with hardware constraints, therefore exploiting the hardware-aware scaling paradigm.

## Acknowledgments

The present research has been partially funded under the EU H2020 project MARVEL (project number: 957337).